%% file: main.tex
\pdfoutput=1

\documentclass[11pt,a4paper]{article}

\usepackage[acceptedWithA]{tacl2021v1}

\usepackage[utf8]{inputenc}
\usepackage[T1]{fontenc}
\usepackage{url}          
\usepackage{latexsym}
\usepackage{times}
\usepackage[export]{adjustbox}
\usepackage{bm}
\usepackage{amstext}
\usepackage{amssymb}
\usepackage{array}
\usepackage{bbm}
\usepackage{boldline}
\usepackage{booktabs}
\usepackage{xcolor}
\usepackage{enumitem}
\usepackage{hyperref}
\usepackage{multirow}
\usepackage[labelfont=bf, format=plain, justification=justified, singlelinecheck=false,font=small]{caption}
\usepackage{subcaption}
\usepackage{tikz}
\usepackage{xspace}
\usepackage{xargs}
\usepackage[colorinlistoftodos,prependcaption,textsize=tiny]{todonotes}
\usetikzlibrary{positioning}
\usepackage{pifont}
\usepackage{makecell}
\usepackage{mathtools}
\usepackage{textcomp}
\usepackage{siunitx}
\usepackage[normalem]{ulem}

\DeclarePairedDelimiterX{\infdivx}[2]{(}{)}{%
  #1\;\delimsize|\delimsize|\;#2%
}
\newcommand{\kld}[2]{\ensuremath{\mathbb{KL}\infdivx{#1}{#2}}\xspace}

\newcolumntype{L}{>{$}l<{$}}
\newcolumntype{C}{>{$}c<{$}}

\newcommand{\cut}[1]{}

\newcommand{\xmark}{\ding{55}}
\newcommand{\cmark}{\ding{51}}

\newcommand{\set}[1]{\mathcal{#1}}
\newcommand\tf[1]{\textbf{#1}}
\newcommand\mbf[1]{\boldsymbol{#1}}

\newcommand{\model}[1]{\textsc{ART}}
\newcommand{\NA}{--}

\title{Questions Are All You Need to Train a Dense Passage Retriever}

\author{
Devendra Singh Sachan$^{1,2}$, Mike Lewis$^{3}$, Dani Yogatama$^{4}$, \\
{\bf Luke Zettlemoyer$^{3,5}$, Joelle Pineau$^{1,2,3}$, Manzil Zaheer$^{4}$} \\
$^{1}$McGill University; 
$^{2}$Mila - Quebec AI Institute;
$^{3}$Meta AI \\
$^{4}$DeepMind;
$^{5}$University of Washington \\
\tt {sachande@mila.quebec}, \{dyogatama, manzilzaheer\}@google.com \\
\tt \{mikelewis,lsz,jpineau\}@meta.com
}

\begin{document}

\maketitle

\input{sections/abstract}

\input{sections/introduction}

\input{sections/methods}

\input{sections/experimental_setup}

\input{sections/results}

\input{sections/related_work}

\input{sections/conclusion}

\input{sections/acknowledgements}

\bibliographystyle{acl_natbib}
\bibliography{main.bib}

\end{document}

%% file: sections/abstract.tex

\begin{abstract}
We introduce \model{}, a new corpus-level autoencoding approach for training dense retrieval models that does not require any labeled training data. 
Dense retrieval is a central challenge for open-domain tasks, such as Open QA, where state-of-the-art methods typically require large supervised datasets with custom hard-negative mining and denoising of positive examples.
\model{}, in contrast, only requires access to unpaired inputs and outputs (e.g.\ questions and potential answer passages). It uses a new passage-retrieval autoencoding scheme, where (1) an input question is used to retrieve a set of evidence passages, and (2) the passages are then used to compute the probability of reconstructing the original question. 
Training for retrieval based on question reconstruction enables effective unsupervised learning of both passage and question encoders, which can be later incorporated into complete Open QA systems without any further finetuning. 
Extensive experiments demonstrate that \model{} obtains state-of-the-art results on multiple QA retrieval benchmarks with only generic initialization from a pre-trained language model, removing the need for labeled data and task-specific losses.\footnote{Our code and model checkpoints are  available at: \url{https://github.com/DevSinghSachan/art}}
\end{abstract}

%% file: sections/introduction.tex

\section{Introduction} 
\label{sec:introduction}

Dense passage retrieval methods~\cite{karpukhin2020dense,xiong2021approximate}, initialized with encoders such as BERT~\cite{devlin2019bert} and trained using supervised contrastive losses~\cite{oord2018representation}, have surpassed the performance achieved by previously popular keyword-based approaches like BM25~\cite{Robertson2009bm25}.
Such retrievers are core components in models for open-domain tasks, such as Open QA, where state-of-the-art methods typically require large supervised datasets with custom hard-negative mining and denoising of positive examples. In this paper, we introduce the first unsupervised method, based on a new corpus-level autoencoding approach, that can match or surpass strong supervised performance levels with no labeled training data or task-specific losses.

We propose \model{}: \emph{Autoencoding-based Retriever Training} which only assumes access to sets of unpaired questions and passages.
Given an input question, \model{} first retrieves a small set of possible evidences passages.
It then  \emph{reconstructs the original question} by attending to these passages (see Figure~\ref{fig:model-figure} for an overview).
The key idea in \model{} is to consider the retrieved passages as a noisy representation of the original question and question reconstruction probability as a way of denoising that provides \emph{soft-labels} for how likely each passage is to have been the correct result.

To bootstrap the training of a strong model, it is important to both have a strong initial retrieval model and to be able to compute reliable initial estimates of question reconstruction probability when conditioned on a (retrieved) passage.
Although passage representations from BERT-style models are known to be reasonable retrieval baselines, it is less clear how to do zero-shot question generation. 
We use a generative pre-trained language model (PLM) and prompt it with the passage as input to generate the question tokens using teacher-forcing.
As finetuning of the question-generation PLM is not needed, only the retrieval model, \model{} can use large PLMs and obtain accurate soft-label estimates of which passages are likely to be the highest quality.

The retriever is trained to penalize the divergence of a passage likelihood from its soft-label score.
For example, if the question is "\texttt{Where is the bowling hall of fame located?}" as shown in Figure~\ref{fig:model-figure}, then the training process will boost the retrieval likelihood of the passage "\texttt{Bowling Hall of Fame is located in Arlington,}" as it is relevant and would lead to a higher question reconstruction likelihood, while the likelihood of the passage "\texttt{Hall of Fame is a song by\ ...}" would be penalized as it is irrelevant.
In this manner, the training process encourages correct retrieval results and vice-versa, leading to an iterative improvement in passage retrieval.

Comprehensive experiments on five benchmark QA datasets demonstrate the usefulness of our proposed training approach.
By simply using questions from the training set, \model{} outperforms models like DPR by an average of 5 points absolute in top-20 and 4 points absolute in top-100 accuracy. 
We also train using all the questions contained in the Natural Questions (NQ) dataset~\cite{Kwiatkowski2019natural} and find that even with a mix of answerable and unanswerable questions, \model{} achieves strong  generalization on out-of-distribution datasets due to relying on PLM. Our analysis further reveals that \model{} is highly sample efficient, outperforming BM25 and DPR with just 100 and 1000 questions, respectively, on the NQ-Open dataset, and that scaling up to larger retriever models consistently  improves performance.

%% file: sections/methods.tex

\input{figures/model-figure}
\section{Method}  \label{sec:modeling-components}

\subsection{Problem Definition}
We focus on open-domain retrieval, where given a question $\boldsymbol{q}$, the task is to select a small set of matching passages (i.e.\ 20 or 100) from a large collection of evidence passages $\set{D}=\{\mbf{d}_1,\ldots, \mbf{d}_m\}$.
Our goal is to train a retriever in a \emph{zero-shot manner}, \emph{i.e.}, without using question-passage pairs, such that it retrieves relevant passages to answer the question.
Our proposed approach consists of two core modeling components (\S\ref{model:dual_encoder}, \S\ref{model:lm_scorer}) and a novel training method (\S\ref{subsec:training_method}).

\subsection{Dual Encoder Retriever} \label{model:dual_encoder}
For the retriever, we use the dual-encoder model~\cite{bromley1994signature} which consists of two encoders, where
\begin{itemize}
	\item one encoder computes the question embedding $f_q(\boldsymbol{q}; \Phi_q):\mathcal{X}\mapsto\mathbb{R}^d$, and
	\item the other encoder computes the passage embedding $f_d(\mbf{d}; \Phi_d):\mathcal{X}\mapsto\mathbb{R}^d$.
\end{itemize}
Here, $\mathcal{X}=\mathbb{V}^n$ denotes the universal set of text sequences, $\mathbb{V}$ denotes the vocabulary consisting of discrete tokens, and $\mathbb{R}^d$ denotes the (latent) embedding space.
We assume that both the question and passage embeddings lie in the same latent space.  
The \emph{retrieval score} for a question-passage pair $(\boldsymbol{q}, \boldsymbol{d})$ is then defined as the inner product between their respective embeddings,
\begin{equation}
\resizebox{.98\hsize}{!}{$\text{score}(\mbf{q}, \mbf{d}_i; \Phi) = f_q(\mbf{q}; \Phi_q) \cdot f_d(\mbf{d}_i; \Phi_d),\ \forall \mbf{d}_i \in \set{D}$ \label{eq:retreiver-score}}
\end{equation}
where $\Phi=[\Phi_q,\Phi_d]$ denotes the retriever parameters.
We select the top-$K$ passages with maximum inner product scores and denote them as $\set{Z}=\{\mbf{z}_1,\ldots,\mbf{z}_K\}$.\footnote{As the selection operation requires performing inner-product with millions of passage embeddings, this can be efficiently performed on accelerators such as GPUs.}

We use the transformer network~\cite{vaswani2017attention} with BERT tokenization~\cite{devlin2019bert} to model both the encoders.
To obtain the question or passage embedding, we do a forward pass through the transformer and select the last layer hidden state corresponding to the \texttt{[CLS]} token.
As the input passage representation, we use both the passage title and text separated by \texttt{[SEP]} token.

\subsection{Zero-Shot Cross-Attention Scorer} \label{model:lm_scorer}
We obtain an estimate of the \emph{relevance score} for a question-(retrieved) passage pair $(\mbf{q},\mbf{z})$ by using a pre-trained language model (PLM).
In order to do this in a zero-shot manner, we use a large generative PLM to compute the likelihood score of a passage conditioned on the question $p(\mbf{z} \mid \mbf{q})$.

The quantity $p(\mbf{z}\mid\mbf{q})$ can be better approximated by the autoregressive generation of question tokens conditioned on the passage and teacher-forcing~\cite{sachan2022improving}.
More formally, this can be written as
\begin{subequations}
\begin{align}
\log p(\mbf{z} & \mid \mbf{q}; \Theta) \nonumber \\
&= \log p(\mbf{q} \mid \mbf{z}; \Theta) + \log p(\mbf{z}) + c \label{eq:1} \\
 &\propto \frac{1}{|\boldsymbol{q}|}\sum_t \log p(q_t \mid \boldsymbol{q}_{<t}, \boldsymbol{z}; \Theta), \label{eq:2}
\end{align}
\end{subequations}
where $\Theta$ denotes the parameters of the PLM, $c$ is a constant independent of the passage $\mbf{z}$, and $|\boldsymbol{q}|$ denotes the number of question tokens.
Here, Eq.~\ref{eq:1} follows from a simple application of Bayes' rule to $p(\mbf{z} \mid \mbf{q})$ and assuming that the passage prior $p(\mbf{z})$ in Eq.~\ref{eq:2} is uniform for all $\mbf{z} \in \set{Z}$.

We hypothesize that calculating the relevance score using Eq.~\ref{eq:2} would be accurate because it requires performing deep cross-attention involving all the question and passage tokens.
In a large PLM, the cross-attention step is highly expressive, and in combination with teacher-forcing, requires the model to explain every token in the question resulting in a better estimation.

As the input passage representation, we concatenate the passage title and its text. In order to prompt the PLM for question generation, we follow~\citet{sachan2022improving} and append a simple natural language instruction "\textit{Please write a question based on this passage.}" to the passage text.

\subsection{Training Algorithm} \label{subsec:training_method}

For training the model, our only assumption is that a collection of questions ($\set{T}$) and evidence passages ($\set{D}$) are provided as input.
During training, the weights of the retriever are updated while the PLM is not finetuned, \emph{i.e.}, it is used in inference mode.
Our training algorithm consists of five core steps. 
The first four steps are performed at every training iteration while the last step is performed every few hundred iterations.
Figure~\ref{fig:model-figure} presents an illustration of our approach.

\paragraph{Step 1: Top-$K$ Passage Retrieval}
For fast retrieval, we pre-compute the evidence passage embedding using the initial retriever parameters ($\hat{\Phi}_d$).
Given a question $\boldsymbol{q}$, we compute its embedding using the current question encoder parameters $(\Phi_q)$ and then retrieve the top-$K$ passages ($\set{Z}$) according to Eq.~\ref{eq:retreiver-score}.
We then embed these top-$K$ passages using the current passage encoder parameters ($\Phi_d$) and compute \emph{fresh} retriever scores as,
\begin{align*}
\text{score}(\mbf{q}, \mbf{z}_i) = f_q(\mbf{q}; \Phi_q) \cdot f_d(\mbf{z}_i; \Phi_d),\ \forall \mbf{z}_i \in \set{Z}.
\end{align*}

\paragraph*{Step 2: Retriever Likelihood Calculation}
Computing the exact likelihood of the passage conditioned on the question requires normalizing over all the evidence passages
\begin{align*}
p(\boldsymbol{z}_i \mid \boldsymbol{q}, \set{D}; \Phi) = \frac{\exp (\text{score}(\boldsymbol{q}, \boldsymbol{z}_i)/\tau)}{\sum_{j=1}^m \exp (\text{score}(\boldsymbol{q}, \boldsymbol{d}_j)/\tau)},
\end{align*}
where $\tau$ is a temperature hyperparameter. 
Computing this term is intractable, as this would require re-embedding all the evidence passages using $\Phi_d$.
Hence, we define a new distribution to approximate the likelihood of $\boldsymbol{z}_i$ as
\begin{align}
\label{eq:ret-prob}
q(\boldsymbol{z}_i \mid \boldsymbol{q}, \set{Z}; \Phi) = \frac{\exp (\text{score}(\boldsymbol{q}, \boldsymbol{z}_i)/\tau)}{\sum_{j=1}^K \exp (\text{score}(\boldsymbol{q}, \boldsymbol{z}_j)/\tau)},
\end{align}
which we also refer to as the \emph{student distribution}.
We assume that passages beyond the top-$K$ contribute a very small probability mass, so we only sum over all the retrieved passages $\set{Z}$ in the denominator. While this approximation leads to a biased estimate of retrieved passage likelihood, it works well in practice. Computing Eq.~\ref{eq:ret-prob} is tractable as it requires embedding and backpropagating through a much smaller set of passages.

\paragraph{Step 3: PLM Relevance Score Estimation}
We compute the relevance score $\log p(\mbf{z}_i \mid \mbf{q})$ of all the passages in $\set{Z}$ using a large PLM ($\Theta$). 
This requires scoring the question tokens using teacher-forcing conditioned on a passage as described in \S\ref{model:lm_scorer}.
We then define a \emph{teacher distribution} by applying softmax to the relevance scores
\begin{align*}
    \hat{p}(\mbf{z}_i \mid \mbf{q}, \set{Z}) = \frac{\exp({\log p(\mbf{z}_i \mid \mbf{q}; \Theta))}}{\sum_{j=1}^K \exp{(\log p(\mbf{z}_j \mid \mbf{q}; \Theta))}}.
\end{align*}

\paragraph{Step 4: Loss Calculation and Optimization}

We train the retriever ($\Phi$) by minimizing the KL divergence loss between the teacher distribution (obtained by PLM) and the student distribution (computed by retriever).
\begin{align*}
    \mathcal{L}(\Phi) = \frac{1}{|\set{T}|}\sum_{\mbf{q}\in\set{T}} \kld{\hat{p}(\mbf{z}_i \mid \mbf{q}, \set{Z})}{q(\boldsymbol{z}_i \mid \boldsymbol{q}, \set{Z}; \Phi)}
\end{align*}
Intuitively, optimizing the KL divergence pushes the passage likelihood scores of the retriever to match the passage relevance scores from PLM by considering the relevance scores as soft-labels.

\paragraph{Step 5: Updating Evidence Embeddings} 
During training, we update the parameters of both the question encoder ($\Phi_q$) and passage encoder ($\Phi_d$).
Due to this, the pre-computed evidence embeddings that was computed using initial retriever parameters ($\hat{\Phi}_d$) becomes stale, which may affect top-$K$ passage retrieval.
To prevent staleness, we re-compute the evidence passage embeddings using current passage encoder parameters ($\Phi_d$) after every 500 training steps.

\subsection{\model{} as an Autoencoder}

Since our encoder takes as input question $\boldsymbol{q}$ and the PLM scores (or reconstructs) the same question when computing the relevance score, we can consider our training algorithm as an autoencoder with a retrieved passage as the latent variable.

In the generative process, we start with an observed variable $\mathcal{D}$ (the collection of evidence passages), which is the support set for our latent variable. Given an input $\boldsymbol{q}$, we generate an index $i$ and retrieve the passage $\boldsymbol{z}_i$. This index generation and retrieval process is modeled by our dual encoder architecture. Given $\boldsymbol{z}_i$, we decode it back into the question using our PLM.

Recall that our decoder (the PLM) is frozen and its parameters are not updated. However, the signal from the decoder output is used to train parameters of the dual encoder such that the log-likelihood of reconstructing the question $\boldsymbol{q}$ is maximized. In practice, this improves the dual encoder to select the best passage for a given question, since the only way to maximize the objective is by choosing the most relevant $\boldsymbol{z}_i$ given the input $\boldsymbol{q}$.

%% file: figures/model-figure.tex

\begin{figure*}[!t]
\centering
\includegraphics[max width=\textwidth, scale=1.0]{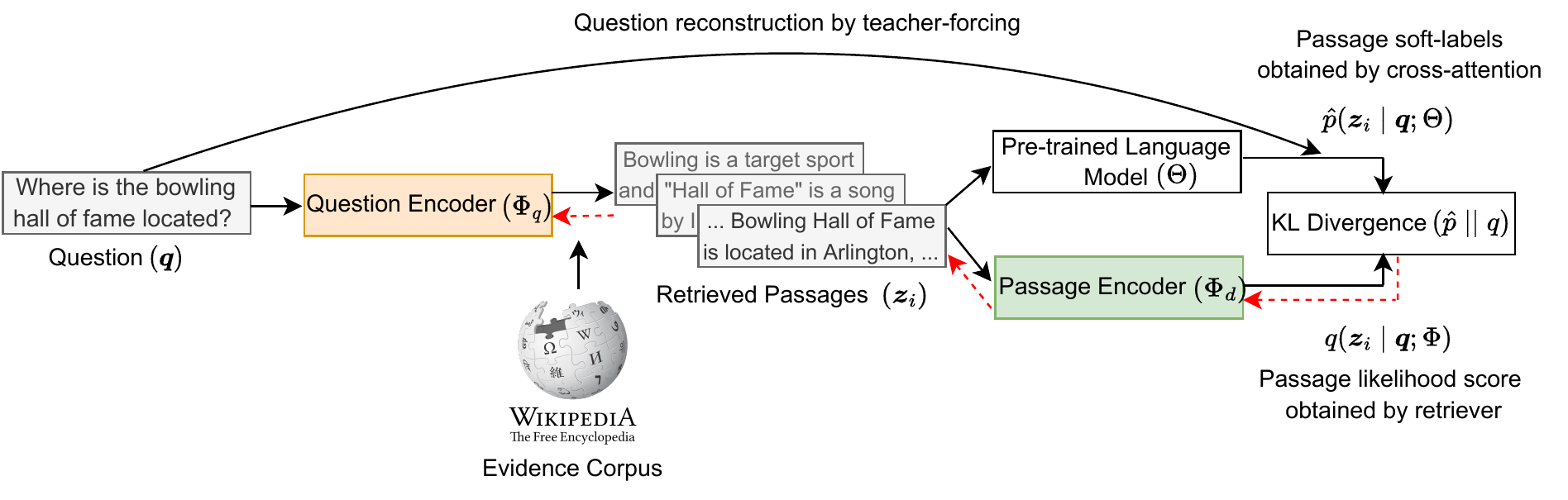}
\caption{
\model{} maximizes the retrieved passage likelihood computed from the dense retriever by considering the language model question reconstruction score conditioned on the passage as a \textit{soft-label}.
Colored blocks indicate trainable parameters. Red arrows show gradient flow during backpropagation.
}
\label{fig:model-figure}
\end{figure*}

%% file: sections/experimental_setup.tex

\section{Experimental Setup}
\label{sec:exp-setup}
In this section, we describe the datasets, evaluation protocol, implementation details, and baseline methods for our passage retrieval experiments.

\subsection{Datasets and Evaluation} \label{subs:datasets}

\input{table/dataset-stats}

\paragraph{Evidence Passages} The evidence corpus includes the preprocessed English Wikipedia dump from December 2018~\cite{karpukhin2020dense}. Following convention, we split an article into non-overlapping segments containing 100 words each resulting in over 21 million passages.
The same evidence is used for both training and evaluation.

\paragraph{Question-Answering Datasets} Following previous work, we use the open-retrieval version of Natural
Questions (NQ-Open;~\citealp{Kwiatkowski2019natural}), TriviaQA~\cite{joshi2017triviaqa}, SQuAD-1.0 (SQuAD-Open;~\citealp{rajpurkar-etal-2016-squad}), WebQuestions (WebQ;~\citealp{berant-etal-2013-semantic}), and EntityQuestions (EQ;~\citealp{sciavolino2021simple}) datasets. 
Table~\ref{tab:dataset_stats} lists their training, development, and test set sizes.

\paragraph{All Questions Datasets}
For our transfer learning experiments, we use all the questions from Natural Questions (henceforth referred to as NQ-Full) and MS MARCO passage ranking~\cite{bajaj2016ms} datasets.
Table~\ref{tab:dataset_stats} lists the number of questions.
The questions in NQ-Full are information-seeking, as they were asked by real users.
Its size is four times that of NQ-Open. 
NQ-Full consists of questions having just long-form of answers such as paragraphs, all the questions in NQ-Open (which have both long-form and short-form answers), questions having yes/no answers, and questions that do not contain the answer or are unanswerable.
For MS MARCO, we use its provided passage collection (around 8.8 million passages in total) as the evidence corpus.

\paragraph{Evaluation}
To evaluate retriever performance, we report the conventional top-$K$ accuracy metric. 
It is the fraction of questions for which at least one passage among the top-$K$ retrieved passages contains a span of words that matches human-annotated answer(s) to the question.

\subsection{Implementation Details}

\paragraph{Model Sizes}
We use BERT base configuration~\cite{devlin2019bert} for the retriever, which consists of 12 layers, 12 attention heads, and 768 embedding dimensions, leading to around 220M trainable parameters.
For the teacher PLM, we use two configurations: (i) T5-XL configuration~\cite{raffel2020t5} consisting of 24 layers, 32 attention heads, and 2048 embedding dimensions, leading to 3B parameters, and (ii) a larger T5-XXL configuration consisting of 11B parameters.

\paragraph{Model Initialization}
We initialize the retriever with unsupervised masked salient spans (MSS) pre-training~\cite{sachan2021end} as it provides an improved zero-shot retrieval over BERT pre-training.\footnote{We use the open-source MSS retriever checkpoint from \url{https://github.com/DevSinghSachan/emdr2}.}
We initialize the cross-attention (or teacher) PLM with the T5-lm-adapted~\cite{lester-etal-2021-power} or instruction-tuned T0~\cite{sanh2022multitask} language models, which have been shown to be effective zero-shot re-rankers for information retrieval tasks~\cite{sachan2022improving}.

\paragraph{Compute Hardware}
We perform training on instances containing 8 or 16 A100 GPUs, each containing 40 GB RAM.

\paragraph{Passage Retrieval}
To perform fast top-$K$ passage retrieval at every training step, we pre-compute the embeddings of all the evidence passages.
Computing embeddings of 21M passages takes roughly 10 minutes on 16 GPUs.
The total size of these embeddings is around 30 GB (768-dimensional vectors in FP16 format). 
For scalable retrieval, we shard these embeddings across all the GPUs and perform exact maximum inner product search using distributed matrix multiplication.

\paragraph{Training Details}
When training with T0 (3B) PLM, for all the datasets except WebQ, we perform training for 10 epochs using Adam with a batch size of 64, 32 retrieved passages, dropout value of 0.1, peak learning rate of \num{2e-5} with warmup and linear scheduling.
Due to the smaller size of WebQ, we train for 20 epochs with a batch size of 16.
When training with the T5-lm-adapted (11B) PLM, we use a smaller batch size of 32 with 16 retrieved passages.
We save the retriever checkpoint every 500 steps and perform model selection by evaluating it on the development set.
We use mixed precision training to train the retriever and perform inference over the PLM using bfloat16 format.
We set the value of the temperature hyperparameter ($\tau$) using cross-validation.

\input{table/main-results}

\subsection{Baselines}  \label{sec:baselines}
We compare \model{} to both unsupervised and supervised models. Unsupervised models train a single retriever using unlabeled text corpus from the Internet while supervised models train a separate retriever for each dataset.
We report the performance numbers from the original papers when the results are available or run their open-source implementations in case the results are not available.

\paragraph*{Unsupervised Models}
These include the popular BM25 algorithm~\cite{Robertson2009bm25} that is based on the sparse bag-of-words representation of text.
Dense models typically use Wikipedia paragraphs to create (pseudo-) query and context pairs to perform contrastive training of the retriever.
These differ in how the negative examples are obtained during contrastive training: they can be from the same batch (ICT;~\citealp{lee-etal-2019-latent,sachan2021end}), or contexts passages from previous batches (Contriever;~\citealp{izacard2021towards}), or by using other passages in the same article (Spider;~\citealp{ram2021learning}). 
Context passages can also be sampled from articles connected via hyperlinks (HLP;~\citealp{zhou2022hyperlink}). 

\paragraph*{Supervised Models}
These consists of approaches that use questions and positive passages to perform contrastive training of the retriever.
To obtain improved performance an additional set of hard-negative passages is often used (DPR;~\citealp{karpukhin2020dense}), iterative mining of negative contexts is done using model weights (ANCE;~\citealp{xiong2021approximate}), or the retriever is first initialized with ICT or MSS pre-training followed by DPR-style finetuning (ICT-DPR / MSS-DPR;~\citealp{sachan2021end}).
The pre-trained retriever can be further trained by ANCE-style mining of hard-negative passages to further improve accuracy (coCondenser;~\citealp{gao-callan-2022-unsupervised}).
Previous methods have also explored finetuning the cross-encoder PLM jointly with the retriever such that cross-encoder provides more accurate training signals to improve retrieval accuracy.
Among them include the approaches of end-to-end training of PLM and retriever which infuses supervision from the annotated answers to a question (\textsc{EMDR}$^2$;~\citealp{sachan2021endtoend}), multi-stage mixed objective distillation approach to jointly train re-ranker~\cite{nogueira2019passage} and retriever (RocketQAv2;~\citealp{ren-etal-2021-rocketqav2}).
A combination of adversarial and distillation-based training of re-ranker and retriever has been shown to obtain state-of-the-art performance (AR2;~\citealp{zhang2022adversarial}).

%% file: table/dataset-stats.tex

\begin{table}[t]
\centering
\footnotesize
\begin{tabular}{l | c c c}
 \toprule
 \textbf{Dataset} & \textbf{Train Questions}  & \textbf{Dev} & \textbf{Test} \\
 \midrule
 \multicolumn{4}{c}{Question-Answering Datasets} \\
 \midrule
 WebQ              & \phantom{00}3,417 & \phantom{00}361 &  \phantom{0}2,032 \\
 NQ-Open                & \phantom{0}79,168 & \phantom{0}8,757 & \phantom{0}3,610 \\
 SQuAD-Open             & \phantom{0}78,713 & \phantom{0}8,886 & 10,570\\
 TriviaQA          & \phantom{0}78,785 & \phantom{0}8,837 & 11,313 \\
 EQ & \NA & 22,068 & 22,075 \\
 \midrule
 \multicolumn{4}{c}{All Questions Datasets} \\
 \midrule
 NQ-Full & 307,373 & \NA & \NA \\
 MS MARCO & 502,939 & \NA & \NA \\
 \bottomrule
 \end{tabular}
\caption{Dataset statistics. During the training process, \model{} only uses the questions while evaluation is performed over the canonical development and test sets.
}
\label{tab:dataset_stats}
\end{table}

%% file: table/main-results.tex

\begin{table*}[t]
\addtolength{\tabcolsep}{-0.25pt}
\footnotesize
\centering
\begin{tabular}{l | l | c c | c c | c c | c c}
\toprule
\tf{Retriever} & \multirow{2}{2.3cm}{\tf{Cross-Attention Language Model}} & \multicolumn{2}{c}{\tf{SQuAD-Open}} & \multicolumn{2}{c}{\tf{TriviaQA}} & \multicolumn{2}{c}{\tf{NQ-Open}} & \multicolumn{2}{c}{\tf{WebQ}} \\
                    & & Top-20 & Top-100 & Top-20 & Top-100 & Top-20 & Top-100 & Top-20 & Top-100 \\
\midrule
\multicolumn{10}{c}{\textit{Unsupervised Approaches (trained using Wikipedia / Internet data)}} \\
\midrule
BERT &  & \phantom{0}5.2 & 13.5 & \phantom{0}7.2 & 17.8 & \phantom{0}9.4 & 20.3 & \phantom{0}3.7 & 12.8  \\
ICT &  & 45.1 & 65.2 & 57.5 & 73.6 & 50.6 & 66.8 & 43.4 & 65.7 \\
MSS & T5$^*$ (220M) & 51.3 & 68.4 & 68.2 & 79.4 & 59.8 & 74.9 & 49.2 & 68.4 \\
BM25 &  & 71.1 & 81.8 & 76.4 & 83.2 & 62.9 & 78.3 & 62.4 & 75.5 \\
Contriever & & 63.4 & 78.2 & 74.2 & 83.2 & 67.8 & 82.1 & 74.9 & 80.1 \\
Spider &  & 61.0 & 76.0 & 75.8 & 83.5 & 68.3 & 81.2 & 65.9 & 79.7 \\
cpt-text S$^\dagger$ & & \NA{} & \NA & 75.1 & 81.7 & 65.5 & 77.2 & \NA & \NA \\
HLP & & \NA & \NA & 76.9 & 84.0 & 70.2 & 82.0 & 66.9 & 80.8 \\
\midrule
\multicolumn{10}{c}{\textit{Supervised Approaches (trained using question-passage aligned data)}} \\
\midrule
DPR &  & 63.2 & 77.2 & 79.4 & 85.0 & 78.4 & 85.4 & 73.2 & 81.4 \\
DPR-Multi$^\ddagger$ & & 51.6 & 67.6 & 78.8 & 84.7 & 79.4 & 86.0 & 75.0 & 82.9 \\
ANCE & & \NA & \NA & 80.3 & 85.3 & 81.9 & 87.5 & \NA & \NA \\
ICT-DPR &  & \NA & \NA & 81.7 & 86.3 & 81.8 & 88.0 & 72.5 & 82.3 \\
MSS-DPR$^\diamond$ &  & 73.1 & 84.5 & 81.8 & 86.6 & 82.1 & 87.8 & 76.9 & 84.6 \\
coCondenser &  & \NA & \NA & 83.2 & 87.3 & 84.3 & 89.0 & \NA & \NA \\
RocketQAv2 & ERNIE$^*$ (110M) & \NA & \NA & \NA & \NA & 83.7 & 89.0 & \NA & \NA \\
\textsc{EMDR}$^2$$^\circ$ & T5$^*$ (220M) & \NA  & \NA & 83.4 & 87.3 & 85.3 & 89.7 & \bf{79.1} & \bf{85.2} \\
AR2 & ERNIE$^*$ (330M) & \NA & \NA & \bf{84.4} & \bf{87.9} & \bf{86.0} & \bf{90.1} & \NA & \NA \\
\midrule
\multicolumn{10}{c}{\textit{Our Approach (trained using questions and Wikipedia text)}} \\
\midrule
\model{} & T5-lm-adapt (11B) & 74.2 & 84.3 & 82.5 & 86.6 & 80.2 & 88.4 & 74.4 & 82.7 \\
\model{}-Multi & T5-lm-adapt (11B) & 72.8 & 83.2 & 82.2 & 86.6 & 81.5 & 88.5 & 74.8	& 83.7 \\
\model{} & T0 (3B) & \underline{75.3} & \underline{85.0} & \underline{82.9} & \underline{87.1} & 81.6 & \underline{89.0} & 75.7 & 84.3 \\
\model{}-Multi & T0 (3B) & 74.7 & 84.5 & \underline{82.9} & 87.0 & \underline{82.0} & 88.9 & \underline{76.6} & \underline{\bf{85.0}} \\
\bottomrule
\end{tabular}
\caption{
Top-20 and top-100 retrieval accuracy on the test set of datasets.
For more details regarding the unsupervised and supervised models, please see \S\ref{sec:baselines} in the text.
Best supervised results are highlighted in bold while best results from the our proposed model (\model{}) are underlined.
\model{} substantially outperforms previous unsupervised models and comes close to or matches the performance of supervised models by just using questions during training.
$^*$ indicates that the cross-attention PLM is finetuned.
$\dagger$ denotes that `cpt-text S' model~\cite{neelakantan2022text} contains around 300M parameters.
$\ddagger$ denotes that DPR-Multi was not trained on SQuAD-Open.
$\diamond$ indicates that the results on SQuAD-Open and WebQ are obtained by finetuning the open-source MSS checkpoint. $\circ$ indicates that $\textsc{EMDR}^2$ results are obtained using their open-source checkpoints.
}
\label{tab:main-results}
\end{table*}

%% file: sections/results.tex

\section{Experiments and Results}  \label{sec:results}

\subsection{Zero-Shot Passage Retrieval}
For the passage retrieval task, we report results on SQuAD-Open, TriviaQA, NQ-Open, and WebQ and train~\model{} under two settings.
In the first setting, we train a separate retriever for each dataset using questions from their training set. 
In the second setting, to examine the robustness of \model{} training to different question types, we train a single retriever by combining the questions from all the four datasets, which we refer to as \model{}-Multi.
For both these settings, we train \model{} using T5-lm-adapted (11B) and T0 (3B) cross-attention PLM scorers.
As our training process does not require annotated passages for a question, we refer to this as \emph{zero-shot passage retrieval}.

Table~\ref{tab:main-results} presents the top-20 and top-100 retrieval accuracy in these settings alongside recent baselines that train a similarly sized retriever (110M). 
All the variants of \model{} achieve substantially better performance than previous unsupervised approaches. 
For example, ART trained with T0 (3B) outperforms the recent Spider and Contriever models by an average of 9 points on top-20 and 6 points on top-100 accuracy.
When comparing to supervised models, despite using just questions, \model{} outperforms strong baselines like DPR and ANCE and is at par or slightly better than pre-trained retrievers like MSS-DPR.
In addition, \model{}-Multi obtains comparable performance to its single dataset version, a considerable advantage in practical applications as a single retriever can be deployed rather than training a custom retriever for each use case.

\model{}'s performance also comes close to the state-of-the-art supervised models like AR2 and \textsc{EMDR}$^2$, especially on the top-100 accuracy but lags behind in the top-20 accuracy. 
In addition to obtaining reasonable performance and not requiring aligned passages for training, \model{}'s training process is much simpler than AR2. 
It also does not require cross-encoder finetuning and is thus faster to train.
As generative language models continue to become more accurate~\cite{Chowdhery2022PaLMSL}, we hypothesize that the performance gap between state-of-the-art supervised models and \model{} would further narrow down.

Our results showcase that both the PLM scorers, T5-lm-adapt (11B) and T0 (3B), achieve strong results on the QA retrieval tasks, with T0 achieving higher performance gains.
This illustrates that the relevance score estimates of candidate passages obtained in the zero-shot cross-attention step are accurate enough to provide strong supervision for retriever training.
We believe that this is a direct consequence of the knowledge stored in the PLM weights.
While T5-lm-adapt's knowledge is obtained by training on unsupervised text corpora, T0 was further finetuned using instruction-prompted datasets of tasks such as summarization, QA, text classification, etc.\footnote{However, we note that T0 was not finetuned on the question generation task and not trained on any of the datasets we have used in this work.
We refer the reader to the original paper for more training details.}
Hence, in addition to learning from instructions, the performance gains from T0 can be attributed to the knowledge infused in its weights by (indirect) supervision from these manually curated datasets.
Instruction-based finetuning is helpful in the case of smaller datasets like WebQ and especially in improving the performance on lower values of top-$K$ accuracy (such as top-20).\footnote{\S\ref{analysis} includes more detailed comparisons of different PLMs as cross-encoders.}

Overall, our results suggest that an \emph{accurate and robust passage retrieval can be achieved by training with questions alone}. 
This presents a considerably more favorable setting than the current approaches which require obtaining positive and hard-negative passages for such questions.
Due to its better performance, we use the T0 (3B) PLM for subsequent experiments unless stated otherwise.

\subsection{Sample Efficiency}
\input{figures/sample-efficiency-plot}
\input{table/nq-transfer-results}
To measure the sample efficiency of \model{}, we train the model by randomly selecting a varying number of questions from NQ-Open training questions and compute the top-$K$ accuracy on its development set.
These results are presented in Figure~\ref{fig:sample-efficiency} and we also include the results of BM25 and DPR for comparison.
We see that performance increases with the increase in questions until about 10k questions, after which the gains become less pronounced.

When trained with just 100 questions, \model{} significantly outperforms BM25 and when trained with 1k questions, it matches DPR performance levels for top-\{50, \ldots, 100\} accuracy.
\emph{This demonstrates that} \model{} \emph{in addition to using just questions is also much more data efficient than} DPR, as it requires almost ten times fewer questions to reach a similar performance.

\subsection{Zero-Shot Out-of-Distribution Transfer}
In the previous experiments, both the training and test sets contained questions that were sampled from the same underlying distribution, a setting that we refer to as \textit{in-distribution training}.
However, obtaining in-domain questions for training is not always feasible in practice.
Instead, a model trained on an existing collection of questions must be evaluated on new datasets, a setting that we refer to as \textit{out-of-distribution} (OOD) transfer. 

We train \model{} using NQ-Open and NQ-Full questions and then evaluate its performance on SQuAD-Open, TriviaQA, WebQ, and EQ datasets.
While it is desirable to train on answerable questions such as the ones included in NQ-Open but this is not always possible, as real user questions are often imprecisely worded or ambiguous. 
Due to this, training on NQ-Full can be considered as a practical testbed for evaluating true OOD generalization as a majority of the questions (51\%) were marked as unanswerable from Wikipedia by human annotators.\footnote{The reasons for question unanswerability can be partly attributed to imprecise Wikipedia article retrieval during the annotation process, ambiguity in information-seeking questions, information required to answer not being localized to a single paragraph, etc~\cite{Kwiatkowski2019natural}.}

Table~\ref{table:nq-transfer-results} presents OOD generalization results on the four QA datasets including the results of DPR and Spider models trained on NQ-Open.\footnote{We also include BM25 results for reference but do not directly compare with them because there is a high lexical overlap between question and passage tokens in the SQuAD-Open and EQ datasets which renders dense retrievers at a disadvantage over BM25, especially in the transfer setting.
}
\model{} trained on NQ-Open always performs significantly better than both DPR and Spider, showing that it is better at generalization than supervised models.
When trained using NQ-Full, \model{} performance further improves by 3 and 0.5-1 points on EQ and other datasets, respectively, over NQ-Open. 
This highlights that in addition to questions annotated as having short answers, \emph{questions annotated with long answers also provide meaningful supervisory signals} and \emph{unanswerable questions do not necessarily degrade performance}.

We also train \model{} using MS MARCO questions and perform OOD evaluation.
Due to the larger size of MS MARCO and a smaller number of evidence passages, we use a batch size of 512 and retrieve 8 passages for training.
Quite surprisingly, it obtains much better performance than previous approaches including BM25 on EQ (more than 10 points gain on top-20 over training \model{} on NQ-Open).
We suspect that this may be due to the similar nature of questions in MS MARCO and EQ.
Further finetuning the pre-trained MS MARCO model on NQ-Full significantly improves performance on WebQ.

\subsection{Scaling Model Size}
\input{table/large-config-retriever}
We examine if scaling up the retriever parameters can offer further performance improvements.
To this end, we train a retriever of BERT-large configuration (24 layers, 16 attention heads, 1024 embedding dimensions) containing around 650M parameters on NQ-Open and TriviaQA.
Results are presented in Table~\ref{tab:large-config-retriever} for both the development and test sets. 
We also include the results of other relevant baselines containing a similar number of trainable parameters.

\emph{By scaling up the retriever size, we see small but consistent improvements in retrieval accuracy across both the datasets}.
Especially on TriviaQA, \model{} matches or exceeds the performance of previous best models.
On NQ-Open, it comes close to the performance of \textsc{EMDR}$^2$~\cite{sachan2021endtoend}, a supervised model trained using thousands of question-answer pairs.

We also attempted to use larger teacher PLMs such as T0 (11B). 
However, our initial experiments did not lead to any further improvements over the T0 (3B) PLM. 
We conjecture that this might be either specific to these QA datasets or that we need to increase the capacity of the teacher PLM even more to observe improvements.
We leave an in-depth analysis of using larger teacher PLMs as part of the future work.

\subsection{Analysis} \label{analysis}

\paragraph{Sensitivity to Retriever Initialization}
\input{figures/effect-of-ret-init-plot}
To examine how the convergence of \model{} training is affected by the initial retriever parameters, we initialize the retriever with (1) BERT weights, (2) ICT weights (as trained in~\citealp{sachan2021end}), and (3) MSS weights, and train using NQ-Open questions.
Figure~\ref{fig:ret-init} displays the top-20 performance on the NQ development set as the training progresses. 
It reveals that \model{} training is not sensitive to the initial retriever parameters as all three initialization schemes converge to similar results.
However, the convergence properties might be different under low-resource settings, an exploration of which we leave for future work.

\paragraph{Effect of the Number of Retrieved Passages}
\input{table/effect-topk-passages.tex}
Table~\ref{tab:effect-topk-passages} quantifies the effect of the number of retrieved passages used during training  on performance.
A smaller number of retrieved passages such as 2 or 4 leads to a somewhat better top-$\{1, 5\}$ accuracy, at the expense of a drop in top-$\{20, 100\}$ accuracy. Retrieving 32 passages offers a reasonable middle ground and beyond that, the top-$K$ retrieval performance tends to drop.

\paragraph{A Closer Inspection of \model{} with Supervised Models}
In order to have a better understanding of the tradeoff between supervised models and \model{}, we examine their top-1 and top-5 accuracy in addition to the commonly reported top-20 and top-100 scores.
Table~\ref{table:fine-grained-cmp} presents these results for \model{} (large) along with supervised models of DPR (large) and EMDR$^2$.
Supervised models achieve much better performance for $\text{top-}K\in\{1,\ldots,5\}$ passages, \emph{i.e.}, these models are more precise. 
This is likely because DPR is trained with hard-negative passages and \textsc{EMDR}$^2$ finetunes PLM using answers resulting in an accurate relevance feedback to the retriever. 
When considering $\text{top-}K\in\{20,\ldots,100\}$ passages, \model{} comes close or matches the performance of \textsc{EMDR}$^2$.
As top-performing models for knowledge-intensive tasks such as open-domain QA rely on a larger set of retrieved passages, such as top-$K$=100~\cite{sachan2022improving}, this justifies the argument to adopt zero-shot \model{} over supervised retrievers.

\input{table/fine-grained-cmp}

\paragraph{Why Training using Passage Retrieval?}
\input{table/passage-effect}
To assess the importance of passages in $\set{Z}$ during the training process, we train the retriever under different settings by varying the passage types.
Specifically, we train with a mix of positive, hard-negative, and uniformly sampled passages.
We also perform in-batch training by defining $\set{Z}$ to be the union of positive and hard-negative passages for all the questions in a batch.
Results in Table~\ref{tab:importance-knn-passages} illustrate that when $\set{Z}$ consists of uniformly sampled passages, it leads to poor performance. 
Including a (gold) positive passage in $\set{Z}$ leads to good performance improvements.
Results further improve with the inclusion of a hard-negative passage in $\set{Z}$.
However, in-batch training leads to a slight drop in performance.
As the gold passages are not always available, our method of selecting the top passages from evidence at every training step can be seen as an approximation to using the gold passages.
With this, \model{} obtains even better results than the previous settings, an improvement by 4 points absolute in the top-20 accuracy.

\paragraph{Impact of Language Model Training Strategy}

\input{table/different-language-model}
\input{table/BEIR}

We examine which PLMs can provide good cross-attention scores during training.
We compare across PLMs trained using three different objectives--- (i) generative denoising of masked spans (T5 series;~\citealp{raffel2020t5}), (ii) further pre-training using autoregressive language modeling objective (T5-lm-adapt series;~\citealp{lester-etal-2021-power}), and (iii) finetuning T5-lm-adapt models on unrelated tasks using  instructions (T0 series;~\citealp{sanh2022multitask}).
Our results in Table~\ref{tab:different-language-model} 
highlight that PLM training methodology and model size can have a large effect on retrieval performance.
T5 base model leads to low scores possibly because pre-training using predicting masked spans is not ideal for question reconstruction. 
However, the accuracy improves with an increase in model size. 
T5-lm-adapt models are more stable and lead to improved performance with the best result achieved by the 11B model.
Instruction finetuned T0 models outperforms the T5-lm-adapt models.
However, scaling up the size of T0 to 11B parameters does not result in meaningful improvements.

\paragraph{Ad-Hoc Retrieval Tasks}
While the previous experiments were conducted on QA datasets, here we examine the robustness of the \model{} model trained using questions to different ad-hoc retrieval tasks.
For this analysis, we evaluate the performance of \model{} on the BEIR benchmark~\cite{thakur2021beir}. It is a heterogeneous collection of many retrieval datasets, with each dataset consisting of test set queries, evidence documents, and gold document annotations.
BEIR spans multiple domains and diverse retrieval tasks presenting a strong challenge suite, especially to the dense retrievers.
We train \model{} using MS MARCO questions and report its nDCG@10 and Recall@100 scores on each dataset.
For comparison, we include the results of three baselines: BM25, Contriever, and DPR trained using NQ-Open.
Our results presented in Table~\ref{tab:beir-benchmark} show strong generalization performance of \model{} as it outperforms DPR and Contriever results. 
\model{} also achieves at par results with the strong BM25 baseline outperforming BM25 on 8 out of the 15 datasets (according to nDCG@10 scores).

%% file: figures/sample-efficiency-plot.tex

\begin{figure}[t!]
\centering
\includegraphics[max width=\linewidth, scale=1.0]{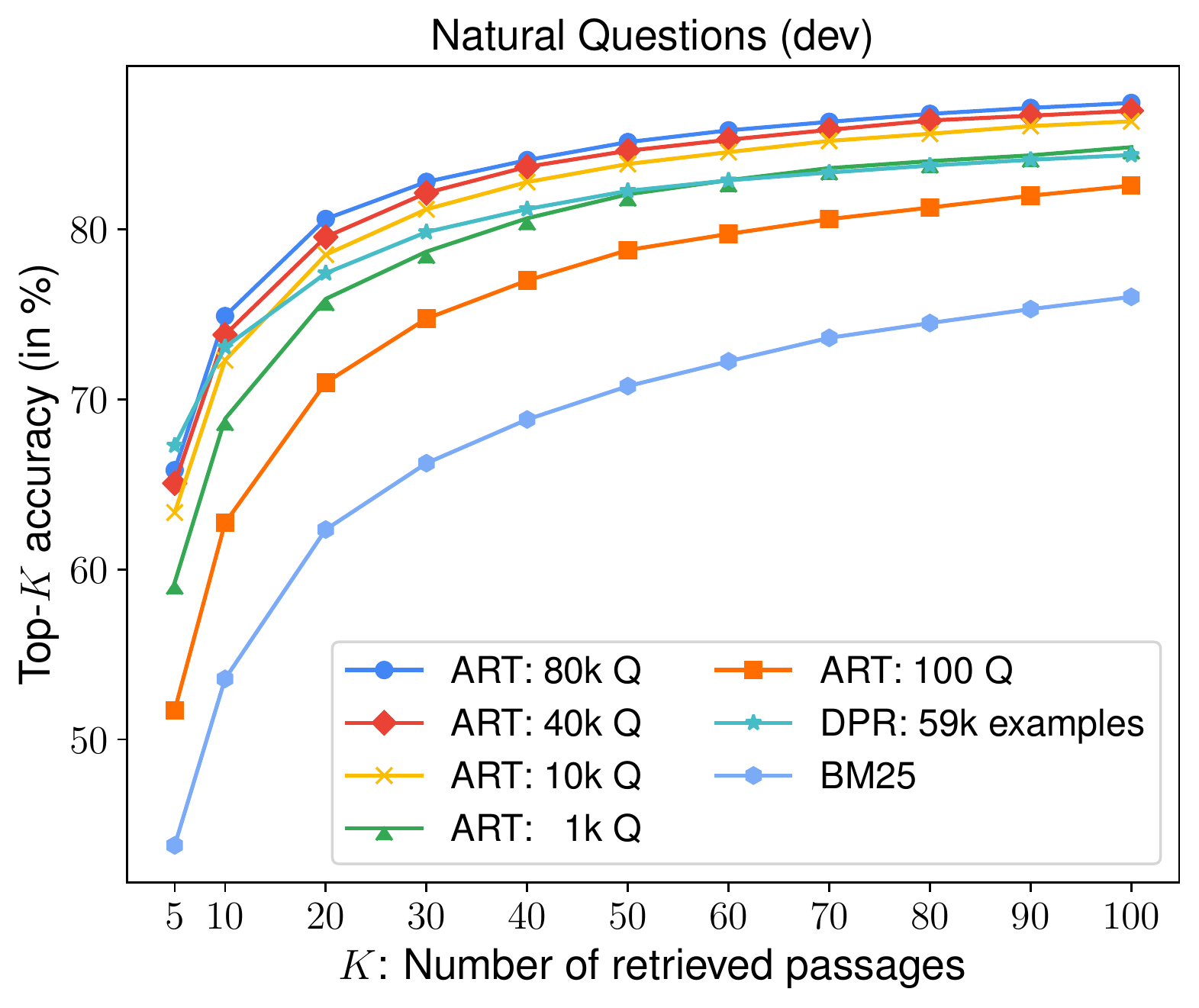}
\caption{
Top-$K$ accuracy as the number of training questions (denoted as `Q' in the legend) is varied.
When trained with 100 questions, \model{} outperforms BM25 and when trained with 1k questions, it matches DPR's performance for top-$K$ > 50 passages, illustrating that \model{} is highly sample efficient.
}
\label{fig:sample-efficiency}
\end{figure}

%% file: table/nq-transfer-results.tex

\begin{table*}[t]
\addtolength{\tabcolsep}{-0.25pt}
\footnotesize
\centering
\begin{tabular}{@{}l  l | c c | c c | c c | c c@{}}
\toprule
\tf{Retriever} & \tf{Training Dataset} & \multicolumn{2}{c}{\tf{SQuAD-Open}} & \multicolumn{2}{c}{\tf{TriviaQA}} & \multicolumn{2}{c}{\tf{WebQ}} & \multicolumn{2}{c}{\tf{EQ}} \\
                              & & Top-20 & Top-100 & Top-20 & Top-100 & Top-20 & Top-100 & Top-20 & Top-100 \\
\midrule
\multicolumn{10}{c}{\textit{Training on answerable questions}} \\
\midrule
BM25 & \NA{}                    & \tf{71.1} & \tf{81.8} & 76.4 & 83.2 & 62.4 & 75.5 & 71.2 & 79.8 \\
DPR$^\dagger$ & NQ-Open         & 48.9 & 65.2 & 69.0 & 78.7 & 68.8 & 78.3 & 49.7 & 63.2 \\
\textsc{EMDR}$^2$ & NQ-Open     & 66.8 & 79.0 & 79.7 & 85.3 & 74.2 & 83.2 & 62.7 & 75.1\\
Spider$^\dagger$ & NQ-Open                & 57.7 & 72.8 & 77.2 & 83.7 & 74.2 & 82.5 & 61.9 & 74.1 \\
\model{} & NQ-Open              & 68.0 & 80.2 & 79.8 & 85.1 & 73.4	& 83.1 & 64.3 & 75.5 \\
\model{} & MS MARCO             & 68.4 & 80.4 & 78.0 & 84.1 & 74.8 & 83.2 & \tf{75.3} & \tf{81.9} \\
\midrule
\multicolumn{10}{c}{\textit{Training on a mix of answerable and unanswerable questions}} \\
\midrule
\model{} & NQ-Full              & 69.4 & \tf{81.1} & 80.3 & \tf{85.7} & 74.3 & 83.9 & 67.8 & 78.3 \\
\model{} & MS MARCO + NQ-Full   & 69.6 & \tf{81.1} & \tf{80.7} & \tf{85.7} & \tf{75.3} & \tf{84.5} & 69.2 & 79.1 \\
\bottomrule
\end{tabular}
\caption{
Top-20 and top-100 retrieval accuracy when evaluating zero-shot out-of-distribution (OOD) generalization of models on the test set of datasets. 
$\dagger$ denotes that these results are from~\citet{ram2021learning}.
For EQ, we report macro-average scores.
\model{} generalizes better than supervised models on OOD evaluation even when trained on all the questions of the Natural Questions dataset which contains a mix of answerable and unanswerable questions.
}
\vspace{-2mm}
\label{table:nq-transfer-results}
\end{table*}

%% file: table/large-config-retriever.tex

\begin{table}[!t]
\addtolength{\tabcolsep}{-1pt}
\footnotesize
\centering
\begin{tabular}{@{}c l c c c c@{}}
\toprule
& \tf{Retriever} & \multicolumn{2}{c}{\tf{NQ-Open}} & \multicolumn{2}{c}{\tf{TriviaQA}} \\
&                & \text{Top-20} & \text{Top-100} & \text{Top-20} & \text{Top-100} \\
\midrule
\multirow{6}{*}{\rotatebox[origin=c]{90}{Dev}} & ICT & 44.2 & 61.0 & 58.8  & 74.4  \\
& DPR      & 79.1 & 85.5 &  81.1 & 85.9 \\
& ICT-DPR  & 81.4 & 87.4 & 82.8 & 86.9 \\
& \textsc{EMDR}$^2$ & \underline{83.1} & \underline{88.0} & \underline{83.7} & \underline{87.4} \\
\cmidrule{3-6}
& \model{}-base  & 80.6	& 87.4 & 83.6 & 87.4 \\
& \model{}-large & \tf{81.0} & \tf{87.8} & \tf{83.7} & \tf{87.5} \\
\midrule
\multirow{6}{*}{\rotatebox[origin=c]{90}{Test}} & ICT & 49.3 & 66.1 & 58.5 & 74.1 \\
& DPR     & 81.0 & 87.2 & 81.4 & 86.0  \\
& ICT-DPR  &  82.6 & 88.3 & 82.9 & 87.1 \\
& \textsc{EMDR}$^2$ & \underline{85.3}	& \underline{89.7} & \underline{83.4} & \underline{87.3} \\
\cmidrule{3-6}
& \model{}-base & 81.6 & \tf{89.0} & 82.9 & 87.1 \\
& \model{}-large & \tf{82.1} & 88.8 & \tf{83.6} & \tf{87.6} \\

\bottomrule
\end{tabular}
 \caption{
Top-20 and top-100 accuracy when training large configuration retriever, which contains around 650M parameters. \textsc{EMDR}$^2$ (base configuration)~\cite{sachan2021endtoend} contains 440M parameters. 
Best supervised results are underlined while the best unsupervised results are highlighted in bold.
}
\label{tab:large-config-retriever}
\end{table}

%% file: figures/effect-of-ret-init-plot.tex

\begin{figure}[t!]
\centering
\includegraphics[max width=\linewidth, scale=0.75]{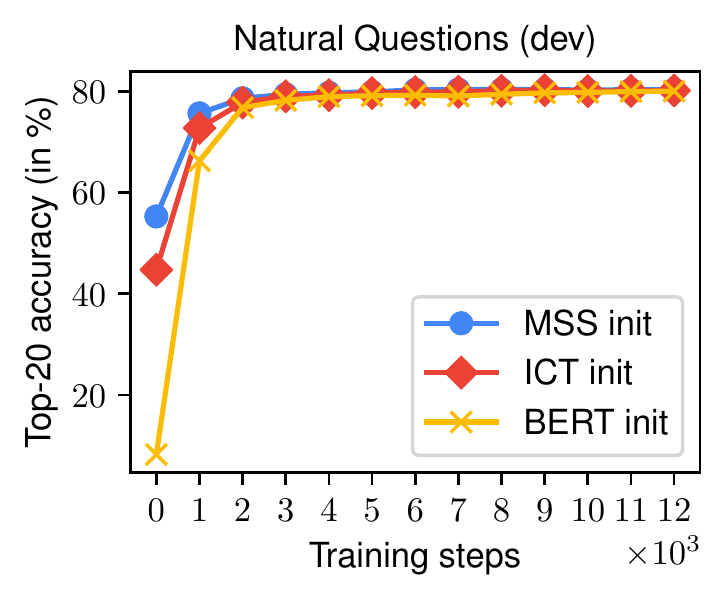}
\vspace{-3mm}
\caption{
Effect of retriever initialization on \model{} training.
The plot reveals that the training process is not sensitive to initial retriever parameters. 
}
\vspace{-2mm}
\label{fig:ret-init}
\end{figure}

%% file: table/effect-topk-passages.tex

\begin{table}[!t]
\footnotesize
\centering
\begin{tabular}{@{}c c c c c@{}}
\toprule
\multirow{2}{1.5cm}{\tf{Retrieved Passages}} & \tf{Top-1} & \tf{Top-5} & \tf{Top-20} & \tf{Top-100} \\
& & \\
\midrule
\phantom{0}32 & 36.7 & 65.8 & 80.6 & 87.4 \\
\midrule
\phantom{00}2 & +2.4  & +0.4  & \textminus0.9 & \textminus0.6 \\
\phantom{00}4 & +1.9  & +0.9  & \textminus0.6 & \textminus0.6 \\
\phantom{00}8 & +0.8  & +0.8  & \textminus0.5 & \textminus0.1 \\
\phantom{0}16 & +0.9  & +0.9  & \textminus0.3 & \textminus0.1 \\
\phantom{0}64 & \textminus 0.5 & \textminus0.7 & \textminus0.3 & 0 \\
128           & \textminus2.3 & \textminus1.7 & \textminus0.8 & \textminus0.2 \\
\bottomrule
\end{tabular}
\caption{
Effect of using a different number of retrieved passages during \model{} training as evaluated on the NQ-Open development set.
For each case, we list the absolute gain or loss in top-$K$ accuracy when compared to the setting utilizing 32 retrieved passages.
}
\label{tab:effect-topk-passages}
\end{table}

%% file: table/fine-grained-cmp.tex

\begin{table}[t]
\footnotesize
\centering
\begin{tabular}{@{}l c c c c@{}}
\toprule
\tf{Retriever} & \text{Top-1} & \text{Top-5} & \text{Top-20} & \text{Top-100} \\
\midrule
\multicolumn{5}{c}{\tf{NQ-Open (dev)}} \\
\midrule
DPR & \tf{50.1} & \tf{69.6} & 79.1 & 85.5 \\
\textsc{EMDR}$^2$ & \tf{55.3} & \tf{74.9} & 83.1 & 88.0 \\
\model{} & 37.6 & 66.8 & 81.0 & 87.8 \\
\midrule
\multicolumn{5}{c}{\tf{TriviaQA (dev)}} \\
\midrule
DPR &  59.6 & 74.4 & 81.1 & 85.9 \\
\textsc{EMDR}$^2$ & \tf{63.7} & 78.0 &	83.7 & 87.4 \\
\model{} & 58.3 & 77.5 & 83.7 & 87.5 \\
\bottomrule
\end{tabular}
 \caption{
Analysis reveals that \model{} (large) can even match the performance of end-to-end trained models like \textsc{EMDR}$^2$ when retrieving a larger number of passages.
However, DPR (large) and \textsc{EMDR}$^2$ still outperform \model{} when retrieving a small number of passages such as $\text{top-}K\in\{1,\ldots,5\}$ (highlighted in bold).
}
\label{table:fine-grained-cmp}
\end{table}

%% file: table/passage-effect.tex

\begin{table}[t]
\footnotesize
\centering
\begin{tabular}{@{}c c c c c c c c@{}}
\toprule
 \bf{P} & \bf{N} & \bf{U} & \bf{IB} & Top-1 & Top-5 & Top-20 & Top-100 \\
\midrule
0 & 0 & 32 & \xmark & \phantom{0}6.0 & 16.6 & 30.8 & 46.7 \\
 1 & 0  & 31 & \xmark & 31.8 & 58.9 & 74.8 & 84.4 \\
 1 & 1 & 30 & \xmark & 33.7 & 61.0 & 76.0 & 85.5 \\
 1 & 1 & \phantom{0}0 & \cmark & 32.6 & 59.5 & 75.1 & 84.9 \\
\midrule
\multicolumn{4}{c}{Top-32 passages} & \bf{36.7} & \bf{65.8} & \bf{80.6} & \bf{87.4} \\
\bottomrule
\end{tabular}
\caption{
Effect of passage types on \model{} training when evaluated on the NQ-Open development set.
P denotes a positive passage, N denotes a hard-negative passage (mined using BM25), U denotes that the passages are randomly sampled from the evidence, and IB denotes in-batch training.
}
\label{tab:importance-knn-passages}
\end{table}

%% file: table/different-language-model.tex
\begin{table}[t]
\footnotesize
\addtolength{\tabcolsep}{-1.pt}
\centering
\begin{tabular}{@{}l| c c c c @{}}
 \toprule
 & \multicolumn{4}{c}{\tf{NQ-Open (dev)}}  \\
 \tf{Language Model} ($\Theta$)           &  Top-1 & Top-5 & Top-20 & Top-100  \\
\midrule
\multicolumn{5}{c}{\textit{Models trained using Denoising Masked Spans}} \\
\midrule
T5-base (250M) &  12.8 & 30.9 & 47.8 & 63.0 \\
T5-xl (3B)         & 25.0 & 53.9 & 74.4 & 85.3 \\
T5-xxl (11B)       &  29.5 & 59.8 & 77.8 & 86.3 \\
\midrule
\multicolumn{5}{c}{\textit{Models trained using Language Modeling Objective}} \\
\midrule
T5-lm-adapt (250M) & 29.4 & 56.6 & 74.4 & 84.7 \\
T5-lm-adapt (800M) & 30.9 & 59.1 & 76.5 & 85.9 \\
T5-lm-adapt (3B)   & 31.8 & 61.0 & 77.9 & 86.5 \\
T5-lm-adapt (11B)  & 32.7 & 62.6 & 78.6 & 87.0 \\
\midrule
\multicolumn{5}{c}{\textit{Model trained using Natural Language Instructions}} \\
\midrule
T0 (3B)  & \bf{36.7} & \bf{65.8} & \bf{80.6} & \bf{87.4} \\
T0 (11B) & 34.3 & 64.5 & 79.8 & 87.2 \\
\bottomrule
\end{tabular}
\caption{
Comparison of different pre-trained language models (PLMs) when used as cross-attention scorers during training (\S\ref{model:lm_scorer}).
T0 (3B) PLM achieves the highest accuracy among the compared PLMs showcasing that training language models using instruction-tuning provides accurate relevance scores.
}
\label{tab:different-language-model}
\end{table}

%% file: table/BEIR.tex

\begin{table*}[t]
\footnotesize
\centering
\begin{tabular}{l c c |c c c c| c c c c}
 \toprule
 \tf{Dataset} & \tf{\#Q} & \tf{\#E} & \multicolumn{4}{c|}{\tf{nDCG@10}} & \multicolumn{4}{c}{\tf{Recall@100}} \\
 \cmidrule{4-11}
 & &  & DPR$^\dagger$ & BM25$^\dagger$ & Contriever & \model{} & DPR$^\dagger$ & BM25$^\dagger$ & Contriever & \model{} \\
\midrule
Scifact & \phantom{00}300        & \phantom{0}5K & 31.8 & \tf{66.5} & 64.9 & 55.2 & 72.7 & 90.8 & \tf{92.6} & 88.0 \\
Scidocs & \phantom{0}1000        & 25K & \phantom{0}7.7 & \tf{15.8} & 14.9 & 14.4 & 21.9 & 35.6 & \tf{36.0} & 32.4 \\
Nfcorpus & \phantom{00}323 & 3.5K & 18.9 & \tf{32.5} & 31.7 & 29.9 & 20.8 & 25.0 & \tf{29.0} & 26.6 \\
FIQA-2018 & \phantom{00}648 & 57K  & 11.2 & 23.6 & 24.5 & \tf{26.5} & 34.2 & 53.9 & \tf{56.2} & 55.4 \\
Trec-covid & \phantom{000}50 & 0.2M & 33.2 & \tf{65.5} & 27.4 & 50.3 & 21.2 & \tf{49.8} & 17.2 & 36.9 \\
Touche-2020 & \phantom{000}49    & 0.4M  & 13.1 & \tf{36.8} & 19.3 & 16.2 & 30.1 & \tf{53.8} & 22.5 & 44.7 \\
NQ & \phantom{0}3452             & 2.7M  & \tf{47.4} & 32.9 & 25.4 & 40.5 & 88.0 & 76.0 & 77.1 & \tf{88.7} \\
MS-Marco & \phantom{0}6980       & 8.8M  & 17.7 & 22.8 & 20.6 & \tf{32.6} & 55.2 & 65.8 & 67.2 & \tf{81.7} \\
HotpotQA & \phantom{0}7405   & 5.2M  & 39.1 & 60.3 & 48.1 & \tf{61.0} & 59.1 & \tf{74.0} & 70.4 & 73.9 \\
ArguAna & \phantom{0}1406   & 8.7K & 17.5 & 31.5 & \tf{37.9} & 32.2 & 75.1 & 94.2 & 90.1 & \tf{95.3} \\
CQADupStack & 13145 & 0.5M & 15.3 & 29.9 & 28.4 & \tf{33.5} & 40.3 & 60.6 & 61.4 & \tf{62.6} \\
Quora & 10000 & 0.5M & 24.8 & 78.9 & 83.5 & \tf{84.2} & 47.0 & 97.3 & 98.7 & \tf{98.8} \\
DBpedia & \phantom{00}400 & 4.6M & 26.3 & 31.3 & 29.2 & \tf{36.3} & 34.9 & 39.8 & 45.3 & \tf{47.2} \\
Fever & \phantom{0}6666 & 5.4M & 56.2 & \tf{75.3} & 68.2 & 72.4 & 84.0 & 93.1 & \tf{93.6} & 93.1 \\
Climate-Fever & \phantom{0}1535 & 5.4M & 14.8 & 21.3 & 15.5 & \tf{21.4} & 39.0 & 43.6 & 44.1 & \tf{47.1} \\
\midrule
\multicolumn{3}{c|}{Average Score} & 25.0 & 41.6 & 36.0 & 40.4 & 48.2 & 63.6 & 60.1 & 64.8 \\
\bottomrule
\end{tabular}
\caption{Results on the BEIR benchmark.
\#Q and \#E denotes the size of the test set and evidence, respectively.
Best scores for each dataset are highlighted in bold.
\model{} is trained using MS MARCO questions.
DPR is trained using NQ-Open.
$\dagger$ denotes that these results are from~\citet{thakur2021beir}.
}
\label{tab:beir-benchmark}
\end{table*}

%% file: sections/related_work.tex

\section{Related Work} \label{sec:related-work}
Our work is based on training a dense retriever using pre-trained language models (PLMs), which we have covered in previous sections. Here, we instead focus on other related approaches.

A popular method to train the dual-encoder retriever is to optimize contrastive loss using in-batch negatives~\cite{gillick-etal-2019-learning} and hard-negatives~\cite{karpukhin2020dense,xiong2021approximate}.
Alternatives to using hard-negatives such as sampling from cached evidence embeddings have also shown to work well in practice~\cite{lindgren2021efficient}.
Multi-vector encoders for questions and passages are more accurate than dual-encoders,~\cite{yi2021sparse,khattab2020colbert,Humeau2020Poly-encoders}, although at the cost of an increased latency and storage requirements.

PLMs have been shown to improve passage rankings as they can perform cross-attention between the question and the retrieved passages~\cite{lin2021pretrained}.
Supervised approaches to re-rank either finetune PLMs using question-passage pairs~\cite{nogueira2020document} or finetune PLMs to generate question conditioned on the passage~\cite{nogueira-dos-santos-etal-2020-beyond} while unsupervised re-rankers are based on zero-shot question scoring~\cite{sachan2022improving}.
The re-ranking process is slow due to the cross-attention step and is bottlenecked by the accuracy of first-stage retrievers.
To address these limitations, cross-attention distillation approaches from the PLM to retriever have been proposed~\cite{qu-etal-2021-rocketqa}.
Such distillation can be performed either in a single end-to-end training step~\cite{guu2020realm,sachan2021endtoend} or in a multi-stage process~\cite{khattab2021relevance,izacard2021distilling}.

An alternative approach to using PLMs is to generate data that can aid retrieval.
The data can be either the title or an answer that provides more information about the question~\cite{mao-etal-2021-generation}.
Generating new questions to augment the training data has also been shown to improve performance~\cite{ma-etal-2021-zero,bonifacio2022inpars,Dai2022DialogIT}.
In comparison, we do not generate new questions but train the retriever using existing questions and PLM feedback.
Data augmentation is likely complementary, and can further improve accuracy.

%% file: sections/conclusion.tex

\section{Conclusions and Future Work} \label{sec:conclusion}
We introduced \model{}, a novel approach to train a dense passage retriever using only questions. 
\model{} does not require question-passage pairs or hard-negative examples for training and yet achieves state-of-the-art results. 
The key to making \model{} work is to optimize the retriever to select relevant passages such that conditioning on them, the question generation likelihood computed using a large pre-trained language model iteratively improves.
Despite requiring much less supervision, \model{} substantially outperforms DPR when evaluated on multiple QA datasets and also generalizes better on out-of-distribution questions.

\model{} presents several directions for future work.
It would be interesting to apply this approach in low-resource retrieval including multi-lingual~\cite{clark2020tydi} and cross-lingual question answering~\cite{asai-etal-2021-xor}. 
Our training framework can also be extended to train cross-modality retrievers such as for image or code search~\cite{li2022competition,neelakantan2022text} using textual queries.
Finally, other directions worth exploring would be to make use of labeled data when available such as by finetuning PLM on passage-question aligned data and to train multi-vector retrievers~\cite{yi2021sparse} with \model{}.

%% file: sections/acknowledgements.tex

\section*{Acknowledgements}
We are grateful to the TACL action editor, Prof.\ Jimmy Lin, and the three anonymous reviewers for providing us valuable feedback and useful suggestions that helped to improve this work.
We would also like to thank Elena Gribovskaya
from DeepMind for providing us valuable comments to improve the paper.

%% file: main.bbl
\begin{thebibliography}{49}
\expandafter\ifx\csname natexlab\endcsname\relax\def\natexlab#1{#1}\fi

\bibitem[{Asai et~al.(2021)Asai, Kasai, Clark, Lee, Choi, and
  Hajishirzi}]{asai-etal-2021-xor}
Akari Asai, Jungo Kasai, Jonathan Clark, Kenton Lee, Eunsol Choi, and Hannaneh
  Hajishirzi. 2021.
\newblock \href {https://doi.org/10.18653/v1/2021.naacl-main.46} {{XOR} {QA}:
  Cross-lingual open-retrieval question answering}.
\newblock In \emph{Proceedings of the 2021 Conference of the North American
  Chapter of the Association for Computational Linguistics: Human Language
  Technologies}.

\bibitem[{Bajaj et~al.(2016)Bajaj, Campos, Craswell, Deng, Gao, Liu, Majumder,
  McNamara, Mitra, Nguyen et~al.}]{bajaj2016ms}
Payal Bajaj, Daniel Campos, Nick Craswell, Li~Deng, Jianfeng Gao, Xiaodong Liu,
  Rangan Majumder, Andrew McNamara, Bhaskar Mitra, Tri Nguyen, et~al. 2016.
\newblock \href {https://arxiv.org/abs/1611.09268} {Ms marco: A human generated
  machine reading comprehension dataset}.
\newblock \emph{arXiv preprint arXiv:1611.09268}.

\bibitem[{Berant et~al.(2013)Berant, Chou, Frostig, and
  Liang}]{berant-etal-2013-semantic}
Jonathan Berant, Andrew Chou, Roy Frostig, and Percy Liang. 2013.
\newblock \href {https://aclanthology.org/D13-1160} {Semantic parsing on
  {F}reebase from question-answer pairs}.
\newblock In \emph{Proceedings of the 2013 Conference on Empirical Methods in
  Natural Language Processing}.

\bibitem[{Bonifacio et~al.(2022)Bonifacio, Abonizio, Fadaee, and
  Nogueira}]{bonifacio2022inpars}
Luiz Bonifacio, Hugo Abonizio, Marzieh Fadaee, and Rodrigo Nogueira. 2022.
\newblock \href {https://doi.org/10.1145/3477495.3531863} {Inpars: Unsupervised
  dataset generation for information retrieval}.
\newblock In \emph{Proceedings of the 45th International ACM SIGIR Conference
  on Research and Development in Information Retrieval}.

\bibitem[{Bromley et~al.(1994)Bromley, Guyon, LeCun, S\"{a}ckinger, and
  Shah}]{bromley1994signature}
Jane Bromley, Isabelle Guyon, Yann LeCun, Eduard S\"{a}ckinger, and Roopak
  Shah. 1994.
\newblock \href
  {https://proceedings.neurips.cc/paper/1993/file/288cc0ff022877bd3df94bc9360b9c5d-Paper.pdf}
  {Signature verification using a "siamese" time delay neural network}.
\newblock In \emph{Advances in Neural Information Processing Systems}.

\bibitem[{Chowdhery et~al.(2022)Chowdhery, Narang, Devlin, Bosma, Mishra,
  Roberts, Barham, Chung, Sutton, Gehrmann, Schuh, Shi, Tsvyashchenko, Maynez,
  Rao, Barnes, Tay, Shazeer, Prabhakaran, Reif, Du, Hutchinson, Pope, Bradbury,
  Austin, Isard, Gur-Ari, Yin, Duke, Levskaya, Ghemawat, Dev, Michalewski,
  Garc{\'i}a, Misra, Robinson, Fedus, Zhou, Ippolito, Luan, Lim, Zoph,
  Spiridonov, Sepassi, Dohan, Agrawal, Omernick, Dai, Pillai, Pellat,
  Lewkowycz, Moreira, Child, Polozov, Lee, Zhou, Wang, Saeta, Diaz, Firat,
  Catasta, Wei, Meier-Hellstern, Eck, Dean, Petrov, and
  Fiedel}]{Chowdhery2022PaLMSL}
Aakanksha Chowdhery, Sharan Narang, Jacob Devlin, Maarten Bosma, Gaurav Mishra,
  Adam Roberts, Paul Barham, Hyung~Won Chung, Charles Sutton, Sebastian
  Gehrmann, Parker Schuh, Kensen Shi, Sasha Tsvyashchenko, Joshua Maynez,
  Abhishek Rao, Parker Barnes, Yi~Tay, Noam~M. Shazeer, Vinodkumar Prabhakaran,
  Emily Reif, Nan Du, Benton~C. Hutchinson, Reiner Pope, James Bradbury, Jacob
  Austin, Michael Isard, Guy Gur-Ari, Pengcheng Yin, Toju Duke, Anselm
  Levskaya, Sanjay Ghemawat, Sunipa Dev, Henryk Michalewski, Xavier Garc{\'i}a,
  Vedant Misra, Kevin Robinson, Liam Fedus, Denny Zhou, Daphne Ippolito, David
  Luan, Hyeontaek Lim, Barret Zoph, Alexander Spiridonov, Ryan Sepassi, David
  Dohan, Shivani Agrawal, Mark Omernick, Andrew~M. Dai,
  Thanumalayan~Sankaranarayana Pillai, Marie Pellat, Aitor Lewkowycz,
  Erica~Oliveira Moreira, Rewon Child, Oleksandr Polozov, Katherine Lee,
  Zongwei Zhou, Xuezhi Wang, Brennan Saeta, Mark Diaz, Orhan Firat, Michele
  Catasta, Jason Wei, Kathleen~S. Meier-Hellstern, Douglas Eck, Jeff Dean, Slav
  Petrov, and Noah Fiedel. 2022.
\newblock \href {https://arxiv.org/abs/2204.02311} {{PaLM}: Scaling language
  modeling with pathways}.
\newblock \emph{ArXiv}, abs/2204.02311.

\bibitem[{Clark et~al.(2020)Clark, Choi, Collins, Garrette, Kwiatkowski,
  Nikolaev, and Palomaki}]{clark2020tydi}
Jonathan~H. Clark, Eunsol Choi, Michael Collins, Dan Garrette, Tom Kwiatkowski,
  Vitaly Nikolaev, and Jennimaria Palomaki. 2020.
\newblock \href {https://doi.org/10.1162/tacl_a_00317} {{T}y{D}i {QA}: A
  benchmark for information-seeking question answering in typologically diverse
  languages}.
\newblock \emph{Transactions of the Association for Computational Linguistics},
  8:454--470.

\bibitem[{Dai et~al.(2022)Dai, Chaganty, Zhao, Amini, Rashid, Green, and
  Guu}]{Dai2022DialogIT}
Zhuyun Dai, Arun~Tejasvi Chaganty, Vincent~Y Zhao, Aida Amini, Qazi~Mamunur
  Rashid, Mike Green, and Kelvin Guu. 2022.
\newblock \href {https://proceedings.mlr.press/v162/dai22a.html} {Dialog
  inpainting: Turning documents into dialogs}.
\newblock In \emph{Proceedings of the 39th International Conference on Machine
  Learning}.

\bibitem[{Devlin et~al.(2019)Devlin, Chang, Lee, and
  Toutanova}]{devlin2019bert}
Jacob Devlin, Ming-Wei Chang, Kenton Lee, and Kristina Toutanova. 2019.
\newblock \href {https://www.aclweb.org/anthology/N19-1423} {{BERT}:
  Pre-training of deep bidirectional transformers for language understanding}.
\newblock In \emph{Proceedings of the 2019 Conference of the North {A}merican
  Chapter of the Association for Computational Linguistics: Human Language
  Technologies, Volume 1 (Long and Short Papers)}.

\bibitem[{Gao and Callan(2022)}]{gao-callan-2022-unsupervised}
Luyu Gao and Jamie Callan. 2022.
\newblock \href {https://doi.org/10.18653/v1/2022.acl-long.203} {Unsupervised
  corpus aware language model pre-training for dense passage retrieval}.
\newblock In \emph{Proceedings of the 60th Annual Meeting of the Association
  for Computational Linguistics (Volume 1: Long Papers)}.

\bibitem[{Gillick et~al.(2019)Gillick, Kulkarni, Lansing, Presta, Baldridge,
  Ie, and Garcia-Olano}]{gillick-etal-2019-learning}
Daniel Gillick, Sayali Kulkarni, Larry Lansing, Alessandro Presta, Jason
  Baldridge, Eugene Ie, and Diego Garcia-Olano. 2019.
\newblock \href {https://doi.org/10.18653/v1/K19-1049} {Learning dense
  representations for entity retrieval}.
\newblock In \emph{Proceedings of the 23rd Conference on Computational Natural
  Language Learning (CoNLL)}.

\bibitem[{Guu et~al.(2020)Guu, Lee, Tung, Pasupat, and Chang}]{guu2020realm}
Kelvin Guu, Kenton Lee, Zora Tung, Panupong Pasupat, and Mingwei Chang. 2020.
\newblock \href {http://proceedings.mlr.press/v119/guu20a.html} {Retrieval
  augmented language model pre-training}.
\newblock In \emph{Proceedings of the 37th International Conference on Machine
  Learning}.

\bibitem[{Humeau et~al.(2020)Humeau, Shuster, Lachaux, and
  Weston}]{Humeau2020Poly-encoders}
Samuel Humeau, Kurt Shuster, Marie-Anne Lachaux, and Jason Weston. 2020.
\newblock \href {https://openreview.net/forum?id=SkxgnnNFvH} {Poly-encoders:
  Architectures and pre-training strategies for fast and accurate
  multi-sentence scoring}.
\newblock In \emph{International Conference on Learning Representations}.

\bibitem[{Izacard et~al.(2022)Izacard, Caron, Hosseini, Riedel, Bojanowski,
  Joulin, and Grave}]{izacard2021towards}
Gautier Izacard, Mathilde Caron, Lucas Hosseini, Sebastian Riedel, Piotr
  Bojanowski, Armand Joulin, and Edouard Grave. 2022.
\newblock \href {https://openreview.net/forum?id=jKN1pXi7b0} {Unsupervised
  dense information retrieval with contrastive learning}.
\newblock \emph{Transactions on Machine Learning Research}.

\bibitem[{Izacard and Grave(2021)}]{izacard2021distilling}
Gautier Izacard and Edouard Grave. 2021.
\newblock \href {https://openreview.net/forum?id=NTEz-6wysdb} {Distilling
  knowledge from reader to retriever for question answering}.
\newblock In \emph{International Conference on Learning Representations}.

\bibitem[{Joshi et~al.(2017)Joshi, Choi, Weld, and
  Zettlemoyer}]{joshi2017triviaqa}
Mandar Joshi, Eunsol Choi, Daniel Weld, and Luke Zettlemoyer. 2017.
\newblock \href {https://www.aclweb.org/anthology/P17-1147} {{T}rivia{QA}: A
  large scale distantly supervised challenge dataset for reading
  comprehension}.
\newblock In \emph{Proceedings of the 55th Annual Meeting of the Association
  for Computational Linguistics (Volume 1: Long Papers)}.

\bibitem[{Karpukhin et~al.(2020)Karpukhin, O{\u{g}}uz, Min, Wu, Edunov, Chen,
  and Yih}]{karpukhin2020dense}
Vladimir Karpukhin, Barlas O{\u{g}}uz, Sewon Min, Ledell Wu, Sergey Edunov,
  Danqi Chen, and Wen-tau Yih. 2020.
\newblock \href {https://www.aclweb.org/anthology/2020.emnlp-main.550} {Dense
  passage retrieval for open-domain question answering}.
\newblock In \emph{Proceedings of the 2020 Conference on Empirical Methods in
  Natural Language Processing (EMNLP)}.

\bibitem[{Khattab et~al.(2021)Khattab, Potts, and
  Zaharia}]{khattab2021relevance}
Omar Khattab, Christopher Potts, and Matei Zaharia. 2021.
\newblock \href {https://doi.org/10.1162/tacl_a_00405} {Relevance-guided
  supervision for openqa with colbert}.
\newblock \emph{Transactions of the Association for Computational Linguistics},
  9:929--944.

\bibitem[{Khattab and Zaharia(2020)}]{khattab2020colbert}
Omar Khattab and Matei Zaharia. 2020.
\newblock \href {https://doi.org/10.1145/3397271.3401075} {Colbert: Efficient
  and effective passage search via contextualized late interaction over bert}.
\newblock In \emph{Proceedings of the 43rd International ACM SIGIR Conference
  on Research and Development in Information Retrieval}.

\bibitem[{Kwiatkowski et~al.(2019)Kwiatkowski, Palomaki, Redfield, Collins,
  Parikh, Alberti, Epstein, Polosukhin, Kelcey, Devlin, Lee, Toutanova, Jones,
  Chang, Dai, Uszkoreit, Le, and Petrov}]{Kwiatkowski2019natural}
Tom Kwiatkowski, Jennimaria Palomaki, Olivia Redfield, Michael Collins, Ankur
  Parikh, Chris Alberti, Danielle Epstein, Illia Polosukhin, Matthew Kelcey,
  Jacob Devlin, Kenton Lee, Kristina~N. Toutanova, Llion Jones, Ming-Wei Chang,
  Andrew Dai, Jakob Uszkoreit, Quoc Le, and Slav Petrov. 2019.
\newblock \href {https://doi.org/10.1162/tacl\_a\_00276} {Natural questions: a
  benchmark for question answering research}.
\newblock \emph{Transactions of the Association of Computational Linguistics},
  7:453--466.

\bibitem[{Lee et~al.(2019)Lee, Chang, and Toutanova}]{lee-etal-2019-latent}
Kenton Lee, Ming-Wei Chang, and Kristina Toutanova. 2019.
\newblock \href {https://www.aclweb.org/anthology/P19-1612} {Latent retrieval
  for weakly supervised open domain question answering}.
\newblock In \emph{Proceedings of the 57th Annual Meeting of the Association
  for Computational Linguistics}.

\bibitem[{Lester et~al.(2021)Lester, Al-Rfou, and
  Constant}]{lester-etal-2021-power}
Brian Lester, Rami Al-Rfou, and Noah Constant. 2021.
\newblock \href {https://doi.org/10.18653/v1/2021.emnlp-main.243} {The power of
  scale for parameter-efficient prompt tuning}.
\newblock In \emph{Proceedings of the 2021 Conference on Empirical Methods in
  Natural Language Processing}.

\bibitem[{Li et~al.(2022)Li, Choi, Chung, Kushman, Schrittwieser, Leblond,
  Eccles, Keeling, Gimeno, Lago, Hubert, Choy, de~Masson~d’Autume,
  Babuschkin, Chen, Huang, Welbl, Gowal, Cherepanov, Molloy, Mankowitz, Robson,
  Kohli, de~Freitas, Kavukcuoglu, and Vinyals}]{li2022competition}
Yujia Li, David Choi, Junyoung Chung, Nate Kushman, Julian Schrittwieser, Rémi
  Leblond, Tom Eccles, James Keeling, Felix Gimeno, Agustin~Dal Lago, Thomas
  Hubert, Peter Choy, Cyprien de~Masson~d’Autume, Igor Babuschkin, Xinyun
  Chen, Po-Sen Huang, Johannes Welbl, Sven Gowal, Alexey Cherepanov, James
  Molloy, Daniel~J. Mankowitz, Esme~Sutherland Robson, Pushmeet Kohli, Nando
  de~Freitas, Koray Kavukcuoglu, and Oriol Vinyals. 2022.
\newblock \href {https://doi.org/10.1126/science.abq1158} {Competition-level
  code generation with alphacode}.
\newblock \emph{Science}, 378(6624):1092--1097.

\bibitem[{Lin et~al.(2021)Lin, Nogueira, and Yates}]{lin2021pretrained}
Jimmy Lin, Rodrigo Nogueira, and Andrew Yates. 2021.
\newblock \href {https://arxiv.org/abs/2010.06467} {{P}retrained {T}ransformers
  for {T}ext {R}anking: {B}ert and {B}eyond}.
\newblock \emph{Synthesis Lectures on Human Language Technologies},
  14(4):1--325.

\bibitem[{Lindgren et~al.(2021)Lindgren, Reddi, Guo, and
  Kumar}]{lindgren2021efficient}
Erik Lindgren, Sashank~J. Reddi, Ruiqi Guo, and Sanjiv Kumar. 2021.
\newblock \href {https://openreview.net/forum?id=824xC-SgWgU} {Efficient
  training of retrieval models using negative cache}.
\newblock In \emph{Advances in Neural Information Processing Systems}.

\bibitem[{Luan et~al.(2021)Luan, Eisenstein, Toutanova, and
  Collins}]{yi2021sparse}
Yi~Luan, Jacob Eisenstein, Kristina Toutanova, and Michael Collins. 2021.
\newblock \href {https://doi.org/10.1162/tacl\_a\_00369} {{Sparse, Dense, and
  Attentional Representations for Text Retrieval}}.
\newblock \emph{Transactions of the Association for Computational Linguistics},
  9:329--345.

\bibitem[{Ma et~al.(2021)Ma, Korotkov, Yang, Hall, and
  McDonald}]{ma-etal-2021-zero}
Ji~Ma, Ivan Korotkov, Yinfei Yang, Keith Hall, and Ryan McDonald. 2021.
\newblock \href {https://doi.org/10.18653/v1/2021.eacl-main.92} {Zero-shot
  neural passage retrieval via domain-targeted synthetic question generation}.
\newblock In \emph{Proceedings of the 16th Conference of the European Chapter
  of the Association for Computational Linguistics: Main Volume}.

\bibitem[{Mao et~al.(2021)Mao, He, Liu, Shen, Gao, Han, and
  Chen}]{mao-etal-2021-generation}
Yuning Mao, Pengcheng He, Xiaodong Liu, Yelong Shen, Jianfeng Gao, Jiawei Han,
  and Weizhu Chen. 2021.
\newblock \href {https://doi.org/10.18653/v1/2021.acl-long.316}
  {Generation-augmented retrieval for open-domain question answering}.
\newblock In \emph{Proceedings of the 59th Annual Meeting of the Association
  for Computational Linguistics and the 11th International Joint Conference on
  Natural Language Processing (Volume 1: Long Papers)}.

\bibitem[{Neelakantan et~al.(2022)Neelakantan, Xu, Puri, Radford, Han, Tworek,
  Yuan, Tezak, Kim, Hallacy et~al.}]{neelakantan2022text}
Arvind Neelakantan, Tao Xu, Raul Puri, Alec Radford, Jesse~Michael Han, Jerry
  Tworek, Qiming Yuan, Nikolas Tezak, Jong~Wook Kim, Chris Hallacy, et~al.
  2022.
\newblock \href {https://arxiv.org/abs/2201.10005} {Text and code embeddings by
  contrastive pre-training}.
\newblock \emph{arXiv preprint arXiv:2201.10005}.

\bibitem[{Nogueira and Cho(2019)}]{nogueira2019passage}
Rodrigo Nogueira and Kyunghyun Cho. 2019.
\newblock \href {https://arxiv.org/abs/1901.04085} {Passage re-ranking with
  {BERT}}.
\newblock \emph{arXiv preprint arXiv:1901.04085}.

\bibitem[{Nogueira et~al.(2020)Nogueira, Jiang, Pradeep, and
  Lin}]{nogueira2020document}
Rodrigo Nogueira, Zhiying Jiang, Ronak Pradeep, and Jimmy Lin. 2020.
\newblock \href {https://doi.org/10.18653/v1/2020.findings-emnlp.63} {Document
  ranking with a pretrained sequence-to-sequence model}.
\newblock In \emph{Findings of the Association for Computational Linguistics:
  EMNLP 2020}.

\bibitem[{Nogueira~dos Santos et~al.(2020)Nogueira~dos Santos, Ma, Nallapati,
  Huang, and Xiang}]{nogueira-dos-santos-etal-2020-beyond}
Cicero Nogueira~dos Santos, Xiaofei Ma, Ramesh Nallapati, Zhiheng Huang, and
  Bing Xiang. 2020.
\newblock \href {https://doi.org/10.18653/v1/2020.emnlp-main.134} {Beyond
  [{CLS}] through ranking by generation}.
\newblock In \emph{Proceedings of the 2020 Conference on Empirical Methods in
  Natural Language Processing (EMNLP)}.

\bibitem[{Oord et~al.(2018)Oord, Li, and Vinyals}]{oord2018representation}
Aaron van~den Oord, Yazhe Li, and Oriol Vinyals. 2018.
\newblock \href {https://arxiv.org/abs/1807.03748} {Representation learning
  with contrastive predictive coding}.
\newblock \emph{arXiv preprint arXiv:1807.03748}.

\bibitem[{Qu et~al.(2021)Qu, Ding, Liu, Liu, Ren, Zhao, Dong, Wu, and
  Wang}]{qu-etal-2021-rocketqa}
Yingqi Qu, Yuchen Ding, Jing Liu, Kai Liu, Ruiyang Ren, Wayne~Xin Zhao, Daxiang
  Dong, Hua Wu, and Haifeng Wang. 2021.
\newblock \href {https://doi.org/10.18653/v1/2021.naacl-main.466}
  {{R}ocket{QA}: An optimized training approach to dense passage retrieval for
  open-domain question answering}.
\newblock In \emph{Proceedings of the 2021 Conference of the North American
  Chapter of the Association for Computational Linguistics: Human Language
  Technologies}.

\bibitem[{Raffel et~al.(2020)Raffel, Shazeer, Roberts, Lee, Narang, Matena,
  Zhou, Li, and Liu}]{raffel2020t5}
Colin Raffel, Noam Shazeer, Adam Roberts, Katherine Lee, Sharan Narang, Michael
  Matena, Yanqi Zhou, Wei Li, and Peter~J. Liu. 2020.
\newblock \href {http://jmlr.org/papers/v21/20-074.html} {Exploring the limits
  of transfer learning with a unified text-to-text transformer}.
\newblock \emph{Journal of Machine Learning Research}, 21(140):1--67.

\bibitem[{Rajpurkar et~al.(2016)Rajpurkar, Zhang, Lopyrev, and
  Liang}]{rajpurkar-etal-2016-squad}
Pranav Rajpurkar, Jian Zhang, Konstantin Lopyrev, and Percy Liang. 2016.
\newblock \href {https://doi.org/10.18653/v1/D16-1264} {{SQ}u{AD}: 100,000+
  questions for machine comprehension of text}.
\newblock In \emph{Proceedings of the 2016 Conference on Empirical Methods in
  Natural Language Processing}.

\bibitem[{Ram et~al.(2022)Ram, Shachaf, Levy, Berant, and
  Globerson}]{ram2021learning}
Ori Ram, Gal Shachaf, Omer Levy, Jonathan Berant, and Amir Globerson. 2022.
\newblock \href {https://aclanthology.org/2022.naacl-main.193} {Learning to
  retrieve passages without supervision}.
\newblock In \emph{Proceedings of the 2022 Conference of the North American
  Chapter of the Association for Computational Linguistics: Human Language
  Technologies}.

\bibitem[{Ren et~al.(2021)Ren, Qu, Liu, Zhao, She, Wu, Wang, and
  Wen}]{ren-etal-2021-rocketqav2}
Ruiyang Ren, Yingqi Qu, Jing Liu, Wayne~Xin Zhao, QiaoQiao She, Hua Wu, Haifeng
  Wang, and Ji-Rong Wen. 2021.
\newblock \href {https://doi.org/10.18653/v1/2021.emnlp-main.224}
  {{R}ocket{QA}v2: A joint training method for dense passage retrieval and
  passage re-ranking}.
\newblock In \emph{Proceedings of the 2021 Conference on Empirical Methods in
  Natural Language Processing}.

\bibitem[{Robertson and Zaragoza(2009)}]{Robertson2009bm25}
Stephen Robertson and Hugo Zaragoza. 2009.
\newblock \href {https://doi.org/10.1561/1500000019} {The {P}robabilistic
  {R}elevance {F}ramework: {BM}25 and {B}eyond}.
\newblock \emph{Foundations and Trends in Information Retrieval}.

\bibitem[{Sachan et~al.(2022)Sachan, Lewis, Joshi, Aghajanyan, Yih, Pineau, and
  Zettlemoyer}]{sachan2022improving}
Devendra~Singh Sachan, Mike Lewis, Mandar Joshi, Armen Aghajanyan, Wen-tau Yih,
  Joelle Pineau, and Luke Zettlemoyer. 2022.
\newblock \href {https://arxiv.org/abs/2204.07496} {Improving passage retrieval
  with zero-shot question generation}.
\newblock In \emph{Proceedings of the 2022 Conference on Empirical Methods in
  Natural Language Processing}.

\bibitem[{Sachan et~al.(2021{\natexlab{a}})Sachan, Patwary, Shoeybi, Kant,
  Ping, Hamilton, and Catanzaro}]{sachan2021end}
Devendra~Singh Sachan, Mostofa Patwary, Mohammad Shoeybi, Neel Kant, Wei Ping,
  William~L Hamilton, and Bryan Catanzaro. 2021{\natexlab{a}}.
\newblock \href {https://aclanthology.org/2021.acl-long.519/} {End-to-end
  training of neural retrievers for open-domain question answering}.
\newblock In \emph{Joint Conference of the 59th Annual Meeting of the
  Association for Computational Linguistics and the 11th International Joint
  Conference on Natural Language Processing (ACL-IJCNLP)}.

\bibitem[{Sachan et~al.(2021{\natexlab{b}})Sachan, Reddy, Hamilton, Dyer, and
  Yogatama}]{sachan2021endtoend}
Devendra~Singh Sachan, Siva Reddy, William~L. Hamilton, Chris Dyer, and Dani
  Yogatama. 2021{\natexlab{b}}.
\newblock \href {https://openreview.net/forum?id=5KWmB6JePx} {End-to-end
  training of multi-document reader and retriever for open-domain question
  answering}.
\newblock In \emph{Advances in Neural Information Processing Systems}.

\bibitem[{Sanh et~al.(2022)Sanh, Webson, Raffel, Bach, Sutawika, Alyafeai,
  Chaffin, Stiegler, Raja, Dey, Bari, Xu, Thakker, Sharma, Szczechla, Kim,
  Chhablani, Nayak, Datta, Chang, Jiang, Wang, Manica, Shen, Yong, Pandey,
  Bawden, Wang, Neeraj, Rozen, Sharma, Santilli, Fevry, Fries, Teehan, Scao,
  Biderman, Gao, Wolf, and Rush}]{sanh2022multitask}
Victor Sanh, Albert Webson, Colin Raffel, Stephen Bach, Lintang Sutawika, Zaid
  Alyafeai, Antoine Chaffin, Arnaud Stiegler, Arun Raja, Manan Dey, M~Saiful
  Bari, Canwen Xu, Urmish Thakker, Shanya~Sharma Sharma, Eliza Szczechla,
  Taewoon Kim, Gunjan Chhablani, Nihal Nayak, Debajyoti Datta, Jonathan Chang,
  Mike Tian-Jian Jiang, Han Wang, Matteo Manica, Sheng Shen, Zheng~Xin Yong,
  Harshit Pandey, Rachel Bawden, Thomas Wang, Trishala Neeraj, Jos Rozen,
  Abheesht Sharma, Andrea Santilli, Thibault Fevry, Jason~Alan Fries, Ryan
  Teehan, Teven~Le Scao, Stella Biderman, Leo Gao, Thomas Wolf, and Alexander~M
  Rush. 2022.
\newblock \href {https://openreview.net/forum?id=9Vrb9D0WI4} {Multitask
  prompted training enables zero-shot task generalization}.
\newblock In \emph{International Conference on Learning Representations}.

\bibitem[{Sciavolino et~al.(2021)Sciavolino, Zhong, Lee, and
  Chen}]{sciavolino2021simple}
Christopher Sciavolino, Zexuan Zhong, Jinhyuk Lee, and Danqi Chen. 2021.
\newblock \href {https://doi.org/10.18653/v1/2021.emnlp-main.496} {Simple
  entity-centric questions challenge dense retrievers}.
\newblock In \emph{Proceedings of the 2021 Conference on Empirical Methods in
  Natural Language Processing}.

\bibitem[{Thakur et~al.(2021)Thakur, Reimers, R{\"u}ckl{\'e}, Srivastava, and
  Gurevych}]{thakur2021beir}
Nandan Thakur, Nils Reimers, Andreas R{\"u}ckl{\'e}, Abhishek Srivastava, and
  Iryna Gurevych. 2021.
\newblock \href {https://openreview.net/forum?id=wCu6T5xFjeJ} {{BEIR}: A
  heterogeneous benchmark for zero-shot evaluation of information retrieval
  models}.
\newblock In \emph{Thirty-fifth Conference on Neural Information Processing
  Systems Datasets and Benchmarks Track (Round 2)}.

\bibitem[{Vaswani et~al.(2017)Vaswani, Shazeer, Parmar, Uszkoreit, Jones,
  Gomez, Kaiser, and Polosukhin}]{vaswani2017attention}
Ashish Vaswani, Noam Shazeer, Niki Parmar, Jakob Uszkoreit, Llion Jones,
  Aidan~N Gomez, {\L}ukasz Kaiser, and Illia Polosukhin. 2017.
\newblock \href
  {https://proceedings.neurips.cc/paper/2017/hash/3f5ee243547dee91fbd053c1c4a845aa-Abstract.html}
  {Attention is all you need}.
\newblock In \emph{Advances in Neural Information Processing Systems}.

\bibitem[{Xiong et~al.(2021)Xiong, Xiong, Li, Tang, Liu, Bennett, Ahmed, and
  Overwijk}]{xiong2021approximate}
Lee Xiong, Chenyan Xiong, Ye~Li, Kwok-Fung Tang, Jialin Liu, Paul~N. Bennett,
  Junaid Ahmed, and Arnold Overwijk. 2021.
\newblock \href {https://openreview.net/forum?id=zeFrfgyZln} {Approximate
  nearest neighbor negative contrastive learning for dense text retrieval}.
\newblock In \emph{International Conference on Learning Representations}.

\bibitem[{Zhang et~al.(2022)Zhang, Gong, Shen, Lv, Duan, and
  Chen}]{zhang2022adversarial}
Hang Zhang, Yeyun Gong, Yelong Shen, Jiancheng Lv, Nan Duan, and Weizhu Chen.
  2022.
\newblock \href {https://openreview.net/forum?id=MR7XubKUFB} {Adversarial
  retriever-ranker for dense text retrieval}.
\newblock In \emph{International Conference on Learning Representations}.

\bibitem[{Zhou et~al.(2022)Zhou, Li, Shang, Luo, Zhan, Hu, Zhang, Jiang, Cao,
  Yu, Jiang, Liu, and Chen}]{zhou2022hyperlink}
Jiawei Zhou, Xiaoguang Li, Lifeng Shang, Lan Luo, Ke~Zhan, Enrui Hu, Xinyu
  Zhang, Hao Jiang, Zhao Cao, Fan Yu, Xin Jiang, Qun Liu, and Lei Chen. 2022.
\newblock \href {https://doi.org/10.18653/v1/2022.acl-long.493}
  {Hyperlink-induced pre-training for passage retrieval in open-domain question
  answering}.
\newblock In \emph{Proceedings of the 60th Annual Meeting of the Association
  for Computational Linguistics (Volume 1: Long Papers)}.

\end{thebibliography}
